# Constrained composite Bayesian optimization for rational synthesis of polymeric particles

Fanjin Wang[1,2,*], Maryam Parhizkar[2], Anthony Harker[3], Mohan Edirisinghe[1]

[1]Department of Mechanical Engineering, University College London, London, WC1E 7JE, United Kingdom

[2]School of Pharmacy, University College London, London, WC1N 1AX, United Kingdom

[3]Department of Physics and Astronomy, University College London, London, WC1E 6BT, United Kingdom

[*]Corresponding author, email: fanjin.wang.20@ucl.ac.uk

**Abstract**

Polymeric nanoparticles have critical roles in tackling healthcare and energy challenges with miniature characteristics. However, tailoring synthesis processes to meet design targets has traditionally depended on domain expertise and trial-and-errors. Modeling strategies, particularly Bayesian optimization, facilitated materials discovery with maximized/minimized properties. Coming from practical demands, this study for the first time integrates constrained composite Bayesian optimization (CCBO) to perform target-value optimization under black-box feasibility constraints for nanoparticle production-by-design. With a synthetic problem that simulates electrospraying, a model nanomanufacturing process, CCBO avoided infeasible conditions and efficiently optimized towards predefined size targets, surpassing baselines and providing comparable decisions as human experts. Laboratory experiments validated CCBO for guided synthesis of poly(lactic-co-glycolic acid) particles with diameters of 300 nm and 3.0 μm *via* electrospraying under minimal initial data. Overall, the CCBO approach presents a versatile and holistic optimization paradigm for next-generation target-driven particle synthesis empowered by artificial intelligence (AI).

## Introduction

Polymeric micro- and nano-particles have received much attention in pharmaceutics, catalysis, and energy applications due to their unique properties at a small scale[1,2]. Diverse design requirements for particles, under the quality-by-design (QbD) framework, have been put forward by specific usages[3]. For example, particles as drug delivery platforms span diverse sizes across hundreds of nanometers for intravenous injection to micrometers for pulmonary administration[4]. However, the optimization of synthesis to meet these design requirements, within any manufacturing technology used, has mainly relied on human expertise and extensive trial-and-error experimentation. Modeling strategies could facilitate the optimization of parameters towards design targets[5,6]. Traditional design of experiment (DoE) strategies can identify dominating factors in the processing parameters and direct towards optimum, but the methodology becomes less effective in high-dimensional problems or complex relationships[7]. For example, orthogonal experiment designs, such as Plackett-Burman and Taguchi methods, typically accommodate up to three levels for each variable[8]. Plus, it is also difficult to incorporate experiment feasibility into DoE optimization frameworks unless analytical descriptions of constraints are available. As a different approach, machine learning (ML) is powerful in modeling complicated relationships[9]. Using ML models as surrogates, adaptive sampling methods design sequential experiments for laboratory evaluation[10]. Bayesian optimization (BO) was developed for efficient optimization of black-box functions, and worked well under small data regime[11,12]. It employs a Gaussian process (GP) as a surrogate model, leveraging its ability to provide both mean and variance estimations for candidate selection. A carefully designed acquisition function is then used to score the candidates to explore uncertain points as well as exploit promising optimal points.

More recently, BO was investigated for materials and drug discovery, assisting the identification of optimal materials properties[13–15]. However, two critical challenges were presented for the application of BO in targeted synthesis of materials. Traditional BO was developed to seek for a global maximum or minimum rather than matching a pre-defined target[16,17], whilst the latter has occurred as a common requirement in functional materials development. Another issue is associated with feasibility constraints in experimentation. The majority of current applications of BO within materials discovery and development[15,18,19] did not incorporate feasibility. Nevertheless, practical concerns could emerge in BO recommendations due to a myriad of reasons in laboratory experiments, such an impossible combination of material compositions, incompatible processing parameters, and limitations from apparatus. Shrinking the boundaries of variables in BO to a more practical region, or imposing known constraints to the optimization process[20,21], could mitigate the issue of generating infeasible experiments. It comes at the cost of sacrificing some of the search space and becomes impossible when the 'practical' region needs to be evaluated through experiments.

Several prior works on constrained and composite BO respectively have explored applications in hyperparameter tuning. In constraint BO, Gramacy and Lee proposed to weight the expected improvement (EI) acquisition function with a modelled probability to enforce preference for feasible candidates[22]. Gardner *et al.* extended this approach to inequality constraints, assuming the feasibility could be derived from a continuous-valued constraint function[23]. More recently, Tian *et al.* proposed boundary exploration method that relaxes acquisition function weights to encourage exploration near constraint boundaries[24]. For composite BO, Uhrenholt and Jensen investigated target value optimization, specifically minimizing a 2-norm, by warping the GP to a noncentral chi-squared distribution[25]. As an improvement, Astudillo and Frazier approached a more general problem of composite BO for arbitrary composite function over the objective function. They transformed the Gaussian posterior in the acquisition function directly with the composited function within the acquisition function[26]. Although these strategies have been rigorously tested on synthetic benchmarks and hyperparameter optimization tasks, they have yet to be integrated into a combinatorial framework to facilitate guided laboratory experiments.

Here, we implement a constrained composite Bayesian optimization (CCBO) pipeline showcasing efficient identification of suitable processing parameters in rational synthesis of polymeric particles. Through introducing a variational inference GP component, the black-box experiment feasibility was modeled and incorporated into BO acquisition function. Composite BO, on the other hand, handles the modeling of experimental parameters and targeting particle size through a composite objective function. Amongst various fabrication techniques of particles, electrospraying was selected as the model technique for its simplicity, versatility, and precision as a popular manufacturing method in drug delivery research[27]. It utilizes electric fields to deform the meniscus of polymer solution to form fine jets which eventually disintegrate into fine droplets. As these droplets travel towards a collector, they further shrink and solidify due to solvent evaporation. Processing parameters in electrospraying such as flow rate, voltage, polymer concentration, and the solvent could be adjusted to tailor product characteristics, although the intertwined impact of these factors could lead to prolonged, if not infeasible, trail-and-errors [28]. With CCBO, we demonstrate its superior performance in target parameter optimization compared with random baseline and conventional BO strategies through both synthetic data and wet-lab experiments of poly(lactic-co-glycolic acid) (PLGA) particles synthesis at multiple size targets.

## Results
**Validating CCBO through synthetic data.** Performance of CCBO was first validated with synthetic experimental data. Before introducing the benchmark results, three configurations of BO pipelines tested in this study are presented (**Fig. 1a**). More details of the implementation can be found in the method section. Briefly, the vanilla BO pipeline followed a traditional BO design where the target to be maximized was the negative squared distance $y_o$. The feasibility component, which leveraged a variational GP for classification, was added to track experimental feasibility. Through factoring in a probability term into the acquisition function, the constrained BO pipeline was able to pick candidates with higher chance of success. CCBO adopted the same feasibility modelling, whilst modifying the objective component. It utilized GP to model the fundamental relationship in experiments between the processing variables $x$ with the size $s$. The negative squared distance function was incorporated in the acquisition function to prioritize candidates for minimizing the distance to the pre-set target. In terms of the synthetic problem, the data was produced by equations simulating electrospray processing. Specifically, the function for determining the size of electrosprayed particles (see **equation (3)**) was inspired by scaling laws proposed for electrospray and experimental observations, where flow rate and polymer concentration (through affecting the viscosity) are both positively correlated to the diameter with voltage having a negative impact[28,29]. Logarithm and power transformations in the function were intended to add complexity in the modeling process to simulate the nonlinear nature of the electrospraying process. The constant for alpha was added to account for the impact of solvents considered in the process. In addition, the feasibility zone, as visualized in **Fig. 1b**, was set to be highly related to the flow rate and the solvent. The rationale was from practical considerations where chloroform, as a highly volatile solvent, would result in clogged nozzle at lower flow rates. N, N-Dimethylacetamide (DMAc), at higher flow rates, would lead to insufficient evaporation of the solvent and produce splashes of droplets on the collector instead of solid particles.

As a benchmark, CCBO, together with random baseline, vanilla BO, and constrained BO only, were performed for 10 iterations. Five initial experiments were included, accounting for successful and failed cases for both solvents. The optimization target was set to 18 μm. Results for other target sizes, including 0.6, 3 and 6 μm, can be found in **Supplementary Fig. 1**. In each iteration, two sets of processing parameters were proposed and subjected to simulation functions to retrieve the synthetic experimental result as well as the feasibility. The regret, defined as the difference between the target and the closest candidate, was recorded after each iteration as a measurement of performance (**Fig. 1c**). After 10 iterations, the random baseline reached 0.8 μm regret. Similarly, the vanilla BO and constrained BO both achieved around 0.4 μm regret. By contrast, the CCBO algorithm rapidly converged to the targeted diameter after only two iterations. Moreover, the area under a curve (AUC) of each strategy, using trapezoidal method, was calculated to quantify the optimization efficiency. CCBO achieved a minimal AUC of 2.47 ± 0.85, significantly lower than random method (19.48 ± 8.12, $p<0.0001$), vanilla BO (18.35 ± 3.86, $p<0.0001$) and constrained BO (16.26 ± 3.73, $p<0.0001$) under one-tailed Mann-Whitney U-test.

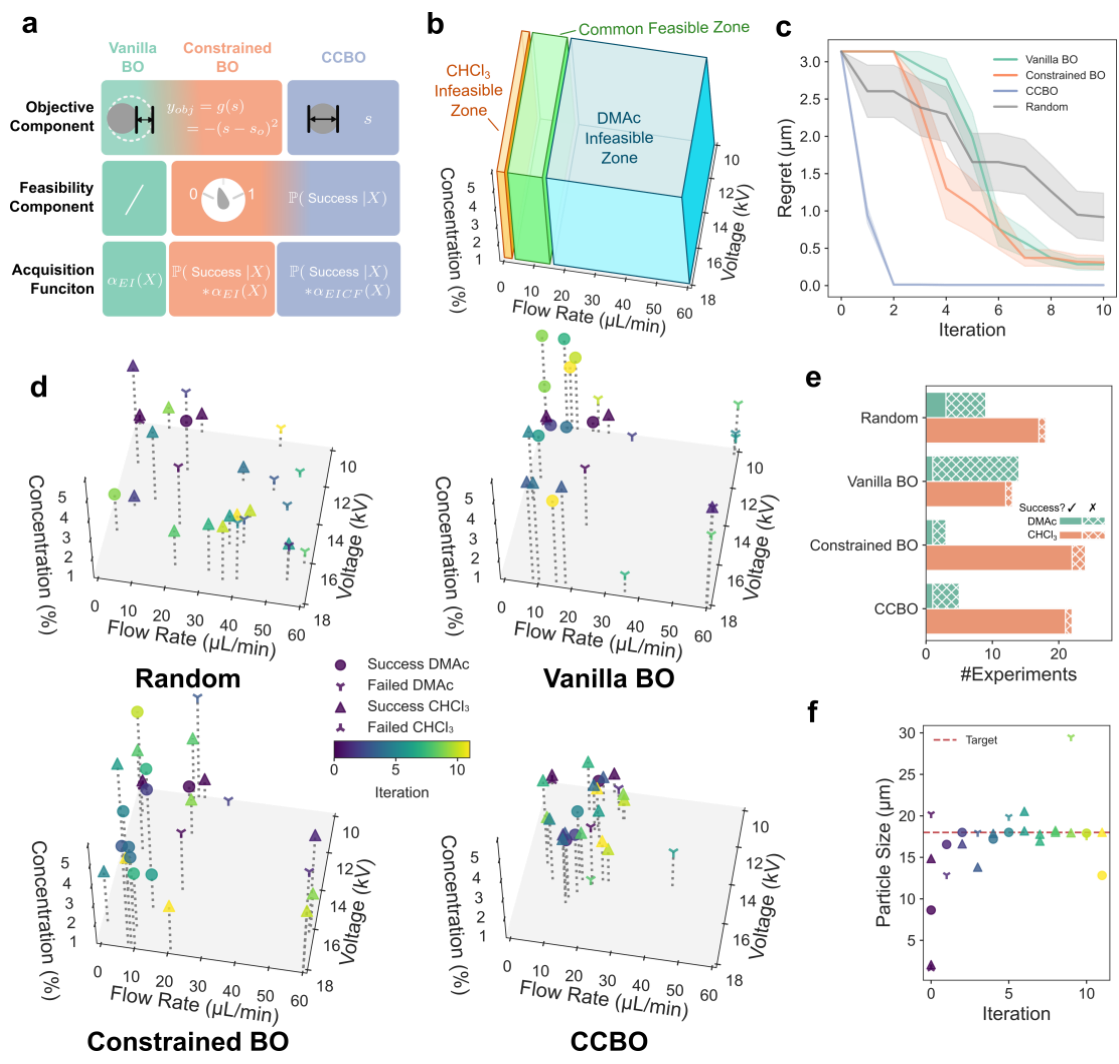

**Fig. 1. Results for CCBO validation with synthetic data. a** An illustration of configurations for vanilla BO, constrained BO, and CCBO. **b** Parameter space visualization for the synthetic data with feasibility zone highlighted for each solvent. **c** Benchmark results of target value optimization with random baseline, vanilla BO, constrained BO, and CCBO. The regret is calculated by the closest distance with respect to the design target, achieved at different iterations of BO. Each benchmark experiment was performed for 10 iterations. Shaded areas indicate standard error from 20 times repetition. **d** Visualization of experimental parameters suggested. Each data point represents one synthetic experiment. The corresponding iteration is coded by color. Symbols represent the solvent used and feasibility of experiment. **e** Comparison of total number of successful (filled bars) and failed experiments (hatched bars) in a typical run of 10 iterations with the four strategies. **f** The particle sizes produced with parameters chosen by CCBO. The iteration is color-coded to the data point and the symbols represent the solvent and feasibility. The target (18 μm) was highlighted as a dashed line.

To understand the recommendation process, the experiments proposed were visualized in **Fig. 1d**. The random baseline sampled uniformly across the experiment space with both solvents, resulting in many failed DMAc experiments due to the flow rate feasibility constraints. Vanilla BO started exploring the boundary conditions in the first few rounds. With an additional model to account for feasibility, the constrained BO algorithm managed to learn the feasible region for DMAc, as reflected by most DMAc experiments being recommended with lower flow rates. This corresponded well to the initial feasible zone visualized in **Fig. 1b**. In addition, the number of failed and successful attempts of each algorithm from the results in **Fig. 1e** were plotted, highlighting the reduction in infeasible experimental conditions with the help of the additional constraint model.

Furthermore, the CCBO strategy was observed to show highly efficient searching in a localized experiment space (**Fig. 1d**). This good performance of CCBO could be explained by its design. The routes taken by vanilla BO and constrained BO were directly minimizing the distance where the surrogate GP was forced to model more complicated results from both the experiment and the superimposed distance function. On the contrary, GP was solely used for modeling the black-box experiment results for CCBO. Our observations with CCBO echoed the findings in composite BO literature: extracting the analytically trackable part from the black-box function can drastically benefit the optimization efficiency[26]. In standard BO, the EI acquisition function assumes Gaussian posterior distribution. However, the posterior of the composite function becomes non-Gaussian after the transformation with a non-linear function. To address this, Astudillo and Frazier suggested leaving the GP to model the black-box function. The composite part was instead incorporated into the acquisition function to transform the Gaussian posterior of the black-box function. This allows more efficient optimization through a closer approximation of posterior distribution in a composite scenario[30]. In our implementation, the composite acquisition function was optimized in the CCBO pipeline with Monte Carlo sampling. Through the benchmark validation, we have shown that vanilla BO or constrained BO alone would not be able to efficiently optimize our design problem, highlighting the importance of the integration of CCBO.

Finally, we compared CCBO to human electrospray users with varying levels of expertise on this synthetic campaign. More experienced users were believed to approach the target more efficiently as they've been equipped with prior knowledge of the parameters' influence and the selection of solvent. All participants (N=14) evaluated the same initial experimental data and suggested experiments to achieve a target particle size of 3 μm. The comparative results are plotted in **Fig. 2a**. In the first iteration, the CCBO strategy was behind intermediate (1-3 years' experience, N=4) and advanced users ($\geq 3$ years' experience, N=4) and performed similarly as beginners (<1 years' experience, N=6). But CCBO soon overtook intermediate users from the second iteration onwards and surpassed advanced user on later iterations. Quantitatively, the AUC was calculated and plotted (**Fig. 2b**) with respect to each strategy or human group, where CCBO (1.40 ± 0.10) owned significantly smaller ($p=0.01$) AUC than beginners (2.62 ± 1.19) under one-tailed Mann-Whitney U-test. There were no significant reductions of AUC with CCBO compared to intermediate (1.60 ± 0.41, $p=0.34$) or advanced (1.03 ± 0.42, $p=0.95$) users. When focusing on overall performance (regret at final iteration), the regret of CCBO strategy was significantly lower than intermediate ($p=0.02$) and beginner ($p<0.0001$) users. Further analysis of parameter selection strategies revealed that advanced users predominantly followed a one-factor-at-a-time (OFAT) approach, resulting in linear adjustment patterns (**Fig. 2c**). Most beginner users and intermediate users attempted to adjust multiple parameters simultaneously. Unlike human participants, CCBO employed more strategic exploration and exploitation, effectively reducing experimental regret by targeting promising regions in the parameter space. Taking together, these findings demonstrated that CCBO could achieve performance comparable to highly experienced participants and navigate complex experimental spaces more effectively than human users. In addition, the performance differences in users with various expertise reflected a successful development of the synthetic problem simulating electrospraying, consolidating our confidence in proceeding to laboratory validation.

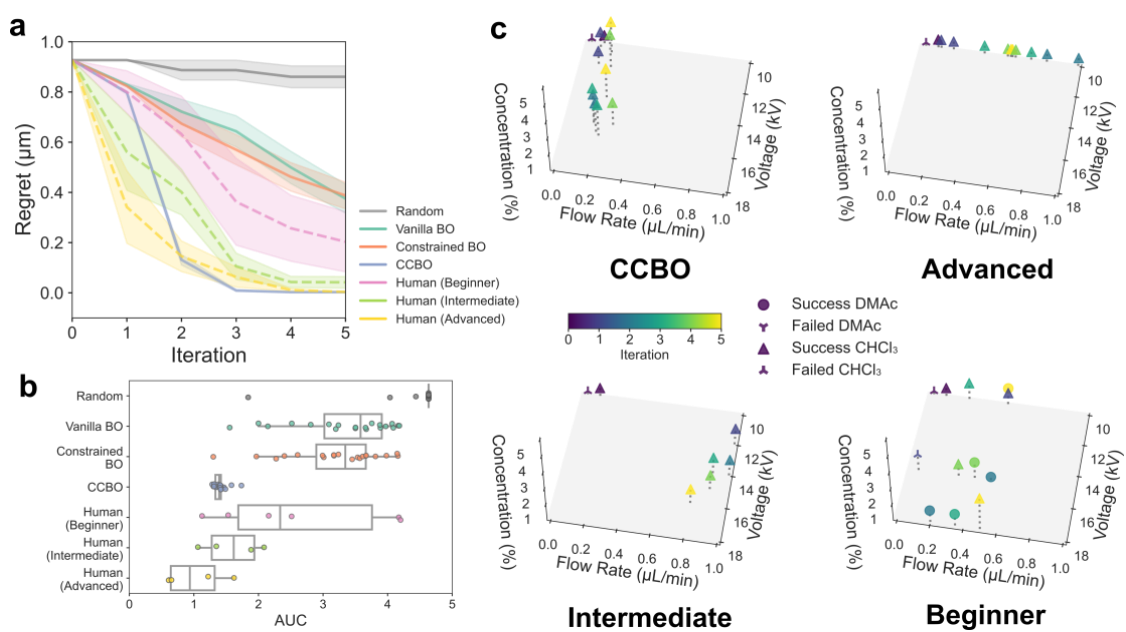

**Fig. 2. Comparing Human and BO performance with synthetic data. a** Benchmark results with BO pipelines (solid lines) for 3.0 μm target in comparison with human users (dashed lines) on synthetic data. The experiment was performed for 5 iterations. Shaded areas for random, vanilla BO, constrained BO, and CCBO indicate standard error from 20 times repetition. Shaded areas for human performance, including beginner (N=6), intermediate (N=4), advanced (N=4), come from standard error of participants respectively. **b** Box plot with scatters on area-under-a-curve (AUC), calculated using trapezoid rule, of the benchmark results from each strategy and human groups. **c** Visualization of experiments selected by CCBO and human participants with various experience levels. Each data point represents one 'synthetic' experiment. The corresponding iteration is coded by color. Symbols represent the solvent used and feasibility of experiment.

**Guiding laboratory electrospraying with CCBO for targeted particle production.** Following the validation of CCBO with synthetic data, it was applied in real-world experiments to guide electrospraying production of micro- and nanoparticles. The initial experiments, generated through a Sobol sequence, were performed to accumulate starting data for BO pipelines (**Table 1**).

**Table 1.** Processing parameters generated through a Sobol sequence and the resulting particle sizes and feasibility (N=3).

| Label | Polymer Concentration (% w/v) | Flow Rate (µL min$^{-1}$) | Voltage (kV) | Solvent | Mean Size (µm) | Feasible? |
|---|---|---|---|---|---|---|
| 0-1 | 2.40 | 1.73 | 14.0 | DMAc | 0.56 | 1 |
| 0-2 | 4.06 | 0.44 | 15.7 | CHCl$_3$ | 1.00 | 0 |
| 0-3 | 2.88 | 49.11 | 11.8 | DMAc | 15.00 | 0 |
| 0-4 | 0.76 | 0.01 | 17.6 | CHCl$_3$ | 1.20 | 0 |
| 0-5 | 0.11 | 10.43 | 14.5 | CHCl$_3$ | 6.26 | 1 |
| 0-6 | 3.55 | 0.06 | 12.8 | DMAc | 0.15 | 1 |
| 0-7 | 4.55 | 2.39 | 16.7 | CHCl$_3$ | 5.24 | 1 |
| 0-8 | 1.88 | 0.21 | 11.0 | DMAc | 1.12 | 1 |

Two particle sizes, 300 nm and 3.0 µm, were set as the design targets based on pharmaceutical interests as drug carriers for intravenous injection and pulmonary delivery[4]. Based on previous reports, the production of PLGA particles with these two particle sizes require distinctive processing parameters involving different solvents and flow rates[28,31]. Thus, the setting of these targets could simulate distinct typical experimental scenarios to challenge BO pipelines. The workflow of targeted particle production under CCBO guidance is illustrated in **Fig. 3a**. With the initial data gathered, CCBO pipeline was implemented to propose two experiments in parallel for laboratory investigation. The selection of two experiments was based on the capacity of laboratory work and to avoid wasting materials and preparation time. After collecting samples and characterization, the results from triplicated experiments were evaluated and compared with the target. The next iteration of BO was performed based on the addition of the new data.

The proposed parameters by CCBO can be visualized with heatmaps in **Fig. 3b**. The heatmap of initial experiments reflected the diverse selections of parameters in Sobol sequence. In total, three iterations of BO were performed for the target of 300 nm and four iterations for 3.0 µm target. The selection of solvents was the most obvious difference for these two targets. Indeed, in previous reports of PLGA particle synthesis, DMAc was a popular solvent due to its high boiling point[32]. From a mechanistic viewpoint, droplets will experience fission due to the competition between Coulombic repulsion and liquid surface tension in an electrospraying process[33]. At the same time, the evaporation of solvents increases the concentration and viscosity of the droplet. As a non-volatile solvent, DMAc allows this fission process to fully develop and thus generates sub-micrometer particles[28]. Chloroform, on the contrary, was preferred in literature to produce lager particles within tens of micrometers range[34]. These practical considerations, normally accumulated through experiences and trial-and-error, were also picked up by the BO pipeline. The recommendations provided by CCBO clearly showed the trend of adopting DMAc for the 300 nm target and chloroform for the 3.0 µm target.

Linking the recommendations to the experiment results (**Fig. 3c**) could provide a more holistic viewpoint of the selection strategy of CCBO. For 300 nm target, the best candidate in initial experiments

(0-8 on **Table 1**) used DMAc with a low polymer concentration, flow rate and voltage to obtain 0.15 μm particles. The recommendations from CCBO pipeline showed exploration of higher concentrations and fine-tuning of the flow rate parameter (**Supplementary Table 1**). Interestingly, the 3-1 and 3-2 experiments both achieved 300 nm particle size with distinct processing parameters, suggesting that the impact from less concentrated polymer solution was compensated by the higher flow rate used for 3-1. Furthermore, the balance of exploration-exploitation from the EI acquisition function was further demonstrated through the experiment series for 3.0 μm target. In the first iteration, CCBO attempted both DMAc and chloroform as the solvent (**Supplementary Table 2**). The second iteration tested the lowest polymer concentration (0.05% w/v), which was shown as the lightest green in the heatmap (**Fig. 3b**). Finally, the recommendation settled down at higher concentration with reduced flow rates to approach the target with fine-tuning from exploitation. It was also observed from the SEM images (**Fig. 3d**) that the experiment 1-2 for 3.0 μm target managed to produce 2.69 μm particles with rough and polydisperse characteristic from a low polymer concentration (0.36% w/v) sprayed at a high flow rate of 3.65 μL min$^{-1}$. The final experiments 4-2 suggested 4.02% w/v solution sprayed at 1.08 μL min$^{-1}$ (**Supplementary Table 2**) to obtain 3.29 μm diameter particles. This result again highlighted the ability to achieve similar particle size through balancing polymer concentration and flow rate, together with adjusting other parameters. The SEM images of the final iteration experiments have shown satisfactory particle production at targeting sizes.

Overall, we have verified the performance of CCBO in the automatic identification of the experiment feasibility region and the rapid convergence to design targets through synthetic data validation. The comparison with human experts demonstrated CCBO's competitive performance. The rational exploration of experiment space outperformed the instinct-driven OFAT trial and error by humans. The wet-lab experiments, as a further step, consolidated CCBO's potential in real-world applications for guided particle synthesis within a few iterations.

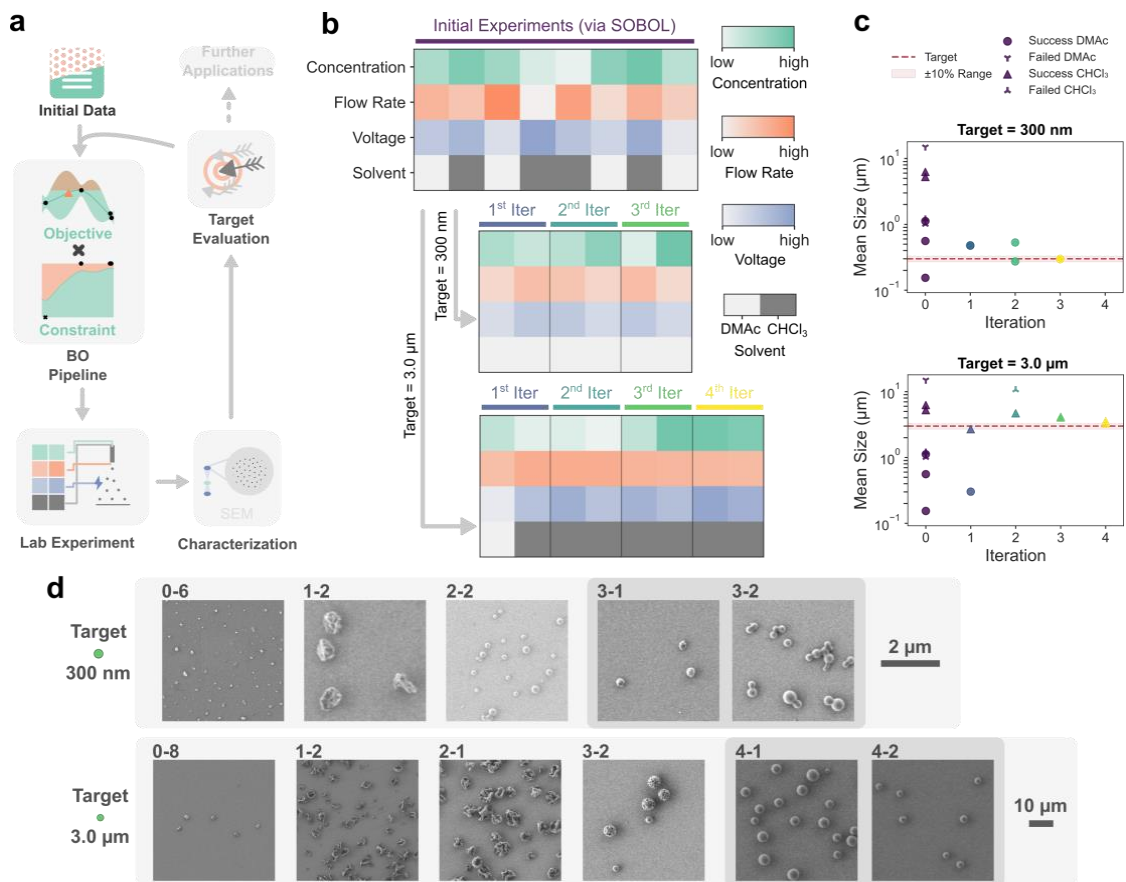

**Fig. 3. Guiding electrospray experiments with CCBO. a** A schematic diagram representing the experiment process with the integration of CCBO. **b** Heatmaps visualizing the processing parameters used for (top) initial experiments, (middle) 300 nm target, and (bottom) 3.0 μm target. The initial experiments were generated with Sobol sequence and the targeted experiment series were suggested by CCBO pipeline. **c** Experiment results of particles generated with electrospraying under parameters proposed for (top) 300 nm and (bottom) 3.0 μm target. Each data point represents the mean of triplicated laboratory experiments. Symbols represent the solvent used and feasibility of experiment. **d** SEM images of particles produced at different iterations for (top) 300 nm and (bottom) 3.0 μm target.

## Discussion

The present work demonstrated the application of efficient CCBO pipeline for target value optimization under black-box constraints. The two components in CCBO worked cohesively to address the need for guiding particle synthesis. For target optimization, the composite BO demonstrated strong capacity in modeling under the composited distance function over the underlying, black-boxed electrospray relationship function. On the other hand, the constraint compartment managed to learn and regulate the suggested experiments with a variational Gaussian process. To deal with unknown feasibility boundaries, many current strategies chose to apply active learning for the identification of unknown feasibility regions, followed by running BO pipelines under the established boundaries[35,36]. As an improvement, CCBO was designed for integrating these two individual processes and focus on identifying the feasibility regions around the design target. This could be seen from the initial experiments where the infeasibility caused by the mismatching of high flow rate with a less volatile solvent DMAc (experiment 0-3 on **Supplementary Table 1**) was not further explored because the target only requires experiments in the lower flow rate region. In comparison, with an active learning pipeline, extra experiments would be needed to determine the possible range for DMAc. Thus, the design of CCBO pipeline allows efficient reduction in the number of experiments to save laboratory resources.

In addition, the innate exploration-exploitation trade-off from BO made possible the identification of multiple possible experimental parameters that can achieve the same design target. This is especially helpful when other design considerations coexist. For example, in the validation with the synthetic problem (**Fig. 1f**), CCBO attempted both DMAc and chloroform and paired them with a wide range of other processing parameters to hit the design target in iterations 6 to 10. From the perspective of production rate, a higher flow rate and polymer concentration might be preferred. Similarly, if the sustainability of the solvent is considered, DMAc would be selected over chloroform as a less harsh solvent. On top of the synthetic data, laboratory experiments also managed to find multiple parameters to produce particles with 300 nm or 3.0 μm diameter. These particles exhibited distinctive morphology and polydispersity, demonstrating varying characteristics for their applications. Although not explicitly coded as a multiple-objective optimization problem, these sets of experimental parameters could be presented to the user as alternative choices. In practice, such flexibility allows the researcher to consider product properties, manufacturing metrics, or other aspects in production, without changing the main design target.

Notably, we highlight that the CCBO pipeline can be seamlessly extended to a broader range of processing parameters, such as involving a wider range of solvents through expanding the boundaries conditions at each iteration. Moreover, CCBO could potentially be extended to other particle synthesis systems, such as batch methods and microfluidics, to facilitate the guided design and production of particles. In the past, the resource-demanding nature of experimentation and scarcity of data posed significant challenges and prolonged the workflow of particle synthesis. We are expecting CCBO to empower nanotechnology with a smarter and more efficient paradigm for target-driven design.

## Methods

**Constrained composite Bayesian optimization.** Two components were incorporated in the BO pipeline and were developed under the framework of BoTorch[37] and GPyTorch[38]. The objective component, which tracked the distance (or particle size in the case of CCBO), followed the classical design of BO (see **Supplementary Note 1** for details of handling categorical inputs)[13]. Notably, due to the difficulty in determining the noise level in experiments, we assumed the input data from laboratory experiments, after averaging over triplicates, to be noiseless. In terms of the acquisition function, *q*-

Expected Improvement ($q$EI, or batch EI) as a thoroughly investigated strategy was selected to allow consideration of multiple candidates jointly in each iteration[16]. In its simplistic form where $q$ equals 1, EI acquisition function at a single point $x_0$ can be given by $\alpha_{EI}(x_0) = \mathbb{E}[\max(y_o - f^*, 0)]$, where $y_o \sim \mathcal{N}(\mu(x_0), \sigma^2(x_0))$ with $\mu(x_0)$ and $\sigma^2(x_0)$ being the posterior mean and variance from the Gaussian process at $x_0$, $f^*$ is the current best observation. As the calculation of expectation requires integrating over the posterior, it becomes analytically intractable under a batched scenario where q>1. We followed the strategy in BoTorch where Monte-Carlo sampling was used to approximate the expectation by:

$$\alpha_{\text{qEI}}(X) \approx \frac{1}{N}\sum_{i=1}^{N} \max_{j=1,\dots,q}[\max(y_{o,ij} - g^*, 0)], y_{o,ij} \sim \mathbb{P}(GP(X)|\mathfrak{D})) \qquad (1)$$

where $N$ was the total number of Monte-Carlo sampling, $q$ was the number of candidates to be evaluated in parallel, and $y_{o,ij}$ was sampled through the reparameterization trick from the Gaussian process conditioned on data $\mathfrak{D}$, $g^*$ represents the current closest distance (with respect to the target) achieved. Notably, the data $\mathfrak{D}$ consisted of $\{(x_i, y_{o,i})\}_{i=1}^{n}$ where $y_{o,i} = g(s_i) = -(s_i - s_o)^2$ with $s_o$ representing the target value. Under such configurations, this vanilla BO pipeline could help identify suitable experiment variables $X$ that can maximize this negative distance measure $y_o$.

Furthermore, the feasibility component was introduced to learn black-box constraints in the experiment. Here, a variational Gaussian process was implemented for the binary classification of experimental success or failure[38]. The details for variational inference for Gaussian classification were described in previous publications[39]. Briefly, the latent Gaussian process is further wrapped with a Probit regression to limit the output between 0 and 1, for the purpose of approximating a Bernoulli posterior. For our latent Gaussian process, it followed the same constant mean prior and kernel functions to incorporate mixed inputs. To incorporate feasibility modelling in the Bayesian optimization process, we followed the strategy proposed earlier[22] to extract the posterior probability as a scaling factor in the acquisition function: $\alpha_{\text{qEIcon}}(X) = \mathbb{P}(y_c = 1|X) * \alpha_{\text{qEI}}(X)$. Incorporating this factor in the acquisition function allowed the suppression of the value of experiments that are potentially infeasible, creating our constrained BO pipeline.

Both the vanilla and constrained BO pipelines had the Gaussian process modeling $y_o$ and attempted to minimize this distance. As a different strategy, composite BO used a Gaussian process to directly model the particle size $s$. The composite part, namely the negative squared distance function $g$, was separated from the input data. Instead, the distance function was directly applied to the Gaussian posterior in the acquisition function[30]:

$$\alpha_{\text{qEICF}}(X) \approx \frac{1}{N}\sum_{i=1}^{N}\max_{j=1,\ldots,q}\left[\max\left(g(s_{ij})-g^{*},0\right)\right], s_{ij}\sim\mathbb{P}(GP(X)|\mathfrak{D}')) \qquad (2)$$

where $s_{ij}$ was sampled through the reparameterization trick and $\mathfrak{D}' = \{(x_i, s_i)\}_{i=1}^{n}$. When coupling the composite acquisition function $\alpha_{\text{qEICF}}$ with the constraint probability, we have the acquisition function for CCBO: $\alpha_{\text{qEICFcon}}(X) = \mathbb{P}(y_c = 1|X) * \alpha_{\text{qEICF}}(X)$.

In the present work, the Monte-Carlo sampling number $N$ was 512 and $q$ was fixed to 2 throughout all BO pipelines. All input $X$ were normalized to unit cubes, and the flow rate variable was transformed to logarithm before normalization. The outcomes of the objective component, including the distance variable, $y_o$, in vanilla BO and constrained BO, as well as the particle size variable, $s$, in CCBO, were standardized to zero mean and unit variance. The outcomes of the feasibility component, $y_c$, were rescaled to $\{-1,1\}$.

**Synthetic electrospray data generation.** The synthetic data of electrospray was generated through the following functions:

$$s = 2 * \frac{\sqrt{QC}}{\log(U)} + \alpha + 0.4 \qquad (3)$$

$$y_c = \begin{cases} 1, & \text{if } \log(Q) * \alpha + 1.4 > 0 \\ 0, & \text{otherwise.} \end{cases} \qquad (4)$$

where $s$ is the particle size (μm), $Q$ is the flow rate (μL min$^{-1}$), $c$ is the concentration of the polymer solution (% w/v), $U$ is the applied voltage (kV). The $\alpha$ is a constant depending on the solvent (CHCl$_3$: 1, DMAc: 0).

**Validating BO with synthetic data.** The targeting particle size $s_o$ was arbitrarily set to be 0.6, 3.0, 6.0 and 18.0 μm to validate BO performance. In each run, three BO pipelines and the random baseline were performed for 10 iterations with the starting data listed on **Table 2**. The outcomes of experiments were calculated by synthetic **equations (3) and (4)** from the corresponding experimental variables. Each run was repeated 20 times to account for variations. The comparison between human and BO followed similar settings used previously. The starting data (**Table 2**) was first shown to the participants (N=14) with varying experience in electrospraying, including advanced users with more than 3 years' experience (N=4), intermediate users between 1-3 years' experience (N=4), and beginners with less than 1 years' experience (N=6). At each iteration, two experiments were recommended by participants to optimize towards a 3.0 μm target, followed by revealing experiment results calculated by the synthetic equations. In total, five iterations were performed for human vs BO campaign.

The regret, defined by the closest distance towards the targeting particle size, was plotted at each iteration. The experimental variables proposed in a typical run were visualized on 3D plots with symbols representing solvent and feasibility, and colors encoding the iteration. AUC of each strategy and human participant was calculated based on trapezoid rules for quantitative comparison. One-tailed Mann-Whitey U-tests were performed with alternative hypothesis being CCBO had smaller AUC/regret compared to BO baselines or human groups, respectively.

Table 2. Boundaries of Experiment Variables for BO and the Starting Data for Synthetic Experiments.

| Label | Polymer Concentration (% w/v) | Flow Rate (µL min$^{-1}$) | Voltage (kV) | Solvent |
|---|---|---|---|---|
| Bounds | [0.05-5.00] | [0.01-60.00] | [10.0-18.0] | {CHCl$_3$, DMAc} |
| S-1 | 0.50 | 15.00 | 10.0 | DMAc |
| S-2 | 0.50 | 0.10 | 10.0 | CHCl$_3$ |
| S-3 | 3.00 | 20.00 | 15.0 | DMAc |
| S-4 | 1.00 | 20.00 | 10.0 | CHCl$_3$ |
| S-5 | 0.20 | 0.02 | 10.0 | CHCl$_3$ |

**Guiding laboratory experiments with CCBO.** The boundaries of experimental variables remained the same as the validation with synthetic data. The starting eight experiments were generated through a Sobol sequence within boundaries for each variable. The targeted particle sizes were 300 nm and 3.0 µm based on domain expertise in drug delivery. The two experiments in each iteration were performed in triplicates. The results were fed back to the BO pipeline to obtain the next recommendations. The stopping criterion was set as achieving ±10% to the targeting size.

**Materials.** PLGA (PURASORB PDLG 5004A, 50:50 ratio) was purchased from Corbion (Amsterdam, The Netherlands). Chloroform and DMAc were purchased from Sigma-Aldrich (Gillingham, UK).

**Electrospraying production of particles.** PLGA solutions were prepared by mixing PLGA granules with solvents at ambient temperature with magnetic stirring overnight. The solutions were fed by a syringe pump (Harvard PHD Ultra, Edenbridge, UK) to a 22-gauge needle (outer diameter 0.71 mm) through a capillary. The positive output of a high voltage power supply (Glassman High Voltage Inc., NJ, United States) was connected to the needle through a crocodile clamp and the collection plate was connected to the ground. Before electrospraying, the flow rate and voltage were adjusted to the values recommended by BO. Experiments were conducted at atmospheric pressure. The temperature and humidity in the room were controlled to be 19-22 °C and 40-50%. Particles were collected on a glass slide placed on the collection plate for Scanning Electron Microscopy (SEM) analysis. Zeiss Gemini 360 SEM (Germany) was used under an acceleration voltage of 1.0 kV with an SE2 detector. For each sample, three images were taken randomly at different locations. Images were further analyzed using ImageJ (National Institute of Health, USA). To obtain mean particle size, a hundred particles were randomly measured for their diameters. For infeasible experiments, the diameters of splashes from undried droplets on the collecting glass slides were recorded as a measurement of size.

## Data availability
Benchmark data in support of this study is available at https://github.com/FrankWanger/CCBO.git. The raw experiment data is available on request.

## Code availability
The code required for reproducing the benchmark results, including the implementation of vanilla BO, constrained BO, and CCBO, are available at https://github.com/FrankWanger/CCBO.git.

## Acknowledgements

The author Fanjin Wang would like to thank the Engineering and Physical Sciences Research Council (EPSRC) for supporting his PhD research (EP/R513143/1 and EP/W524335/1). Dr Jakob Zeitler from Matterhorn Studio is thanked for initial discussions. The participants of the human vs. BO campaign are gratefully acknowledged for their time and expertise.

## Author contributions

F.W. conceptualized the study, developed methodology and software, and performed investigation. M.P., A.H., and M.E. helped with methodology. M.P. and M.E. provided resources and supervision. F.W. drafted the original manuscript. M.P., A.H., and M.E. edited and revised the manuscript. All authors approved the final version.

## Competing interests

The authors declare no competing interests.

Supplementary Information for

# Constrained composite Bayesian optimization for rational synthesis of polymeric particles

Fanjin Wang[1,2,*], Maryam Parhizkar[2], Anthony Harker[3], Mohan Edirisinghe[1]

[1]Department of Mechanical Engineering, University College London, London, WC1E 7JE, United Kingdom

[2]School of Pharmacy, University College London, London, WC1N 1AX, United Kingdom

[3]Department of Physics and Astronomy, University College London, London, WC1E 6BT, United Kingdom

[*]Corresponding author, email: fanjin.wang.20@ucl.ac.uk

# Supplementary Notes
## Supplementary Note 1: Surrogate development for CCBO and baseline BO methods.

A Gaussian process with constant mean prior was used as the surrogate function[1]. As the input consists of both continuous (i.e., processing parameters) and categorial (solvents) variables, the covariance module of the Gaussian process adopted the design in BoTorch[2] to combine the categorial and a continuous kernel. More specifically, the mixed kernel was defined by:

$$\begin{aligned}k_{\text{Mixed}}\left((x_{c,1}, x_{d,1}), (x_{c,2}, x_{d,2})\right) \\ = k_{\text{Matérn52}}(x_{c,1}, x_{c,2}) + k_{\text{Hamming}}(x_{d,1}, x_{d,2}) \\ + k_{\text{Matérn52}}(x_{c,1}, x_{c,2}) * k_{\text{Hamming}}(x_{d,1}, x_{d,2}),\end{aligned} \quad (1)$$

where $x_c$ and $x_d$ were the continuous and discrete compartments in the input, respectively. The $k_{\text{Matérn52}}$ was a Matérn kernel with the smoothness parameter $v$ set as 5/2:

$$k_{\text{Matérn52}}(x_{c,1}, x_{c,2}) = \frac{2^{1-v}}{\Gamma(v)} \left(\sqrt{2v}d\right)^v K_v\left(\sqrt{2v}d\right), \quad (2)$$

where $d = (x_{c,1} - x_{c,2})^\top \Theta (x_{c,1} - x_{c,2})$ with $\Theta$ being the length scale parameter, $\Gamma$ was the gamma function, and $K_v$ was a modified Bessel function. The $k_{\text{Hamming}}$ was a categorial kernel based on Hamming distance:

$$k_{\text{Hamming}}(x_{d,1}, x_{d,2}) = e^{-\frac{HD(x_{d,1}, x_{d,2})}{\Theta}}, \quad (3)$$

where $HD$ was the Hamming distance function.

## Supplementary Tables:

**Supplementary Table 1.** Processing parameters generated through CCBO for target 300 nm and the resulting particle sizes and feasibility (N=3).

| Label | Polymer Concentration (% w/v) | Flow Rate (µL min$^{-1}$) | Voltage (kV) | Solvent | Mean Size (µm) | Feasible? |
|---|---|---|---|---|---|---|
| 1-1 | 2.32 | 0.09 | 12.0 | DMAc | 0.48 | 1 |
| 1-2 | 1.33 | 0.74 | 13.9 | DMAc | 0.47 | 1 |
| 2-1 | 1.89 | 0.43 | 13.9 | DMAc | 0.53 | 1 |
| 2-2 | 3.52 | 0.10 | 12.3 | DMAc | 0.27 | 1 |
| 3-1 | 0.58 | 0.84 | 14.0 | DMAc | 0.30 | 1 |
| 3-2 | 4.61 | 0.07 | 12.4 | DMAc | 0.30 | 1 |

**Supplementary Table 2.** Processing parameters generated through CCBO for target 3.0 µm and the resulting particle sizes and feasibility (N=3).

| Label | Polymer Concentration (% w/v) | Flow Rate (µL min$^{-1}$) | Voltage (kV) | Solvent | Mean Size (µm) | Feasible? |
|---|---|---|---|---|---|---|
| 1-1 | 1.66 | 0.80 | 10.7 | DMAc | 0.30 | 1 |
| 1-2 | 0.36 | 3.65 | 14.6 | CHCl$_3$ | 2.69 | 1 |
| 2-1 | 0.57 | 3.74 | 16.5 | CHCl$_3$ | 4.69 | 1 |
| 2-2 | 0.05 | 3.55 | 14.5 | CHCl$_3$ | 10.64 | 0 |
| 3-1 | 1.63 | 1.92 | 16.2 | CHCl$_3$ | 4.14 | 1 |
| 3-2 | 4.51 | 1.38 | 14.9 | CHCl$_3$ | 4.05 | 1 |
| 4-1 | 4.45 | 1.30 | 17.4 | CHCl$_3$ | 3.58 | 1 |
| 4-2 | 4.02 | 1.08 | 16.3 | CHCl$_3$ | 3.29 | 1 |

## Supplementary Figures:

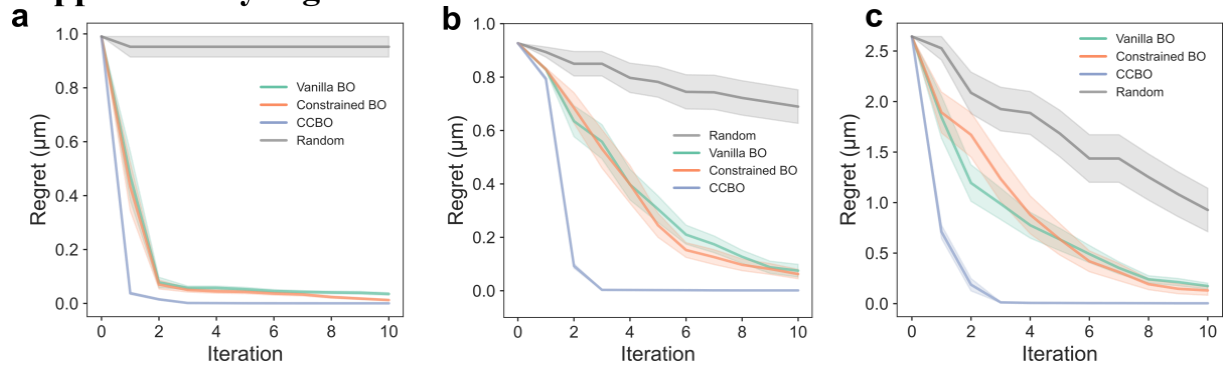

**Supplementary Figure 1.** Benchmark results with random baseline, vanilla BO, constrained only BO, and CCBO for (**a**) 0.6 μm, (**b**) 3.0 μm and (**c**) 6.0 μm targets. Each benchmark experiment was performed for 10 iterations. Shaded areas indicate standard error from 20 times repetition.

## Supplementary References